\documentclass[10pt,twocolumn,letterpaper]{article}

\usepackage{iccv}
\usepackage{times}
\usepackage{epsfig}
\usepackage{graphicx}
\usepackage{amsmath}
\usepackage{amssymb}

\usepackage{pdfpages}
\usepackage[accsupp]{axessibility}
\usepackage[symbol]{footmisc}

\usepackage{multirow}
\usepackage{ctable}
\newcommand\scalemath[2]{\scalebox{#1}{\mbox{\ensuremath{\displaystyle #2}}}}

\DeclareMathOperator*{\argmax}{argmax}
\usepackage[breaklinks=true,bookmarks=false]{hyperref}

\iccvfinalcopy 

\def\iccvPaperID{9951} 

\ificcvfinal\fi

\begin{document}

\title{Complementary Domain Adaptation and Generalization\\ for Unsupervised Continual Domain Shift Learning}

\author{Wonguk Cho$^1$, Jinha Park$^2$, and Taesup Kim$^{1*}$\\
$^1$Graduate School of Data Science, $^2$Department of Electrical and Computer Engineering\\
Seoul National University\\
{\tt\small \{wongukcho, jhpark410, taesup.kim\}@snu.ac.kr}
}
\maketitle
\footnotetext{*Corresponding Author}
\ificcvfinal\thispagestyle{empty}\fi

\begin{abstract}
Continual domain shift poses a significant challenge in real-world applications, particularly in situations where labeled data is not available for new domains. The challenge of acquiring knowledge in this problem setting is referred to as unsupervised continual domain shift learning. Existing methods for domain adaptation and generalization have limitations in addressing this issue, as they focus either on adapting to a specific domain or generalizing to unseen domains, but not both. In this paper, we propose Complementary Domain Adaptation and Generalization (CoDAG), a simple yet effective learning framework that combines domain adaptation and generalization in a complementary manner to achieve three major goals of unsupervised continual domain shift learning: adapting to a current domain, generalizing to unseen domains, and preventing forgetting of previously seen domains. Our approach is model-agnostic, meaning that it is compatible with any existing domain adaptation and generalization algorithms. We evaluate CoDAG on several benchmark datasets and demonstrate that our model outperforms state-of-the-art models in all datasets and evaluation metrics, highlighting its effectiveness and robustness in handling unsupervised continual domain shift learning.
\end{abstract}

\begin{figure}[ht]
\begin{center}
\includegraphics[width=\linewidth]{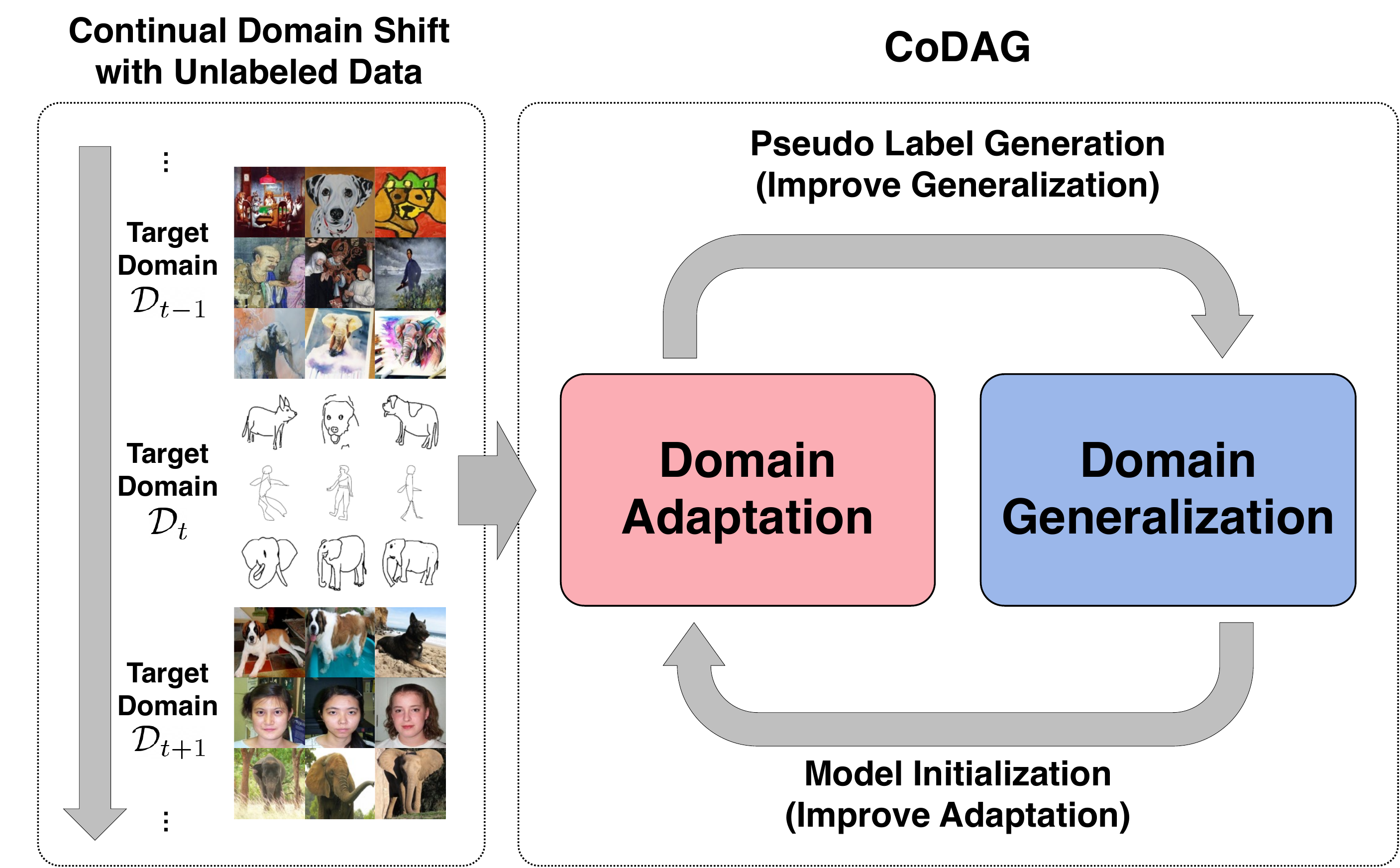}

\end{center}
   \caption{Our proposed Complementary Domain Adaptation and Generalization (CoDAG) framework for unsupervised continual domain shift learning.}
\label{fig:short}
\end{figure}

\section{Introduction}
Machine learning algorithms have found extensive applications in various fields such as image recognition~\cite{he2016deep, krizhevsky2017imagenet}, natural language processing~\cite{bengio2000neural, devlin2018bert}, and autonomous driving~\cite{bojarski2016end, dosovitskiy2017carla}. Typically, these algorithms learn from a training dataset to build a model that performs a target task on new data. The assumption underlying these algorithms is that the training data and the test data are identically and independently distributed (IID)~\cite{hastie2009elements}, drawn from the same distribution that is characterized by an environment or domain. However, this IID assumption is often not valid in real-world scenarios, as the environment in which the model is applied is more likely to change over time than remain fixed. This implies that the model will encounter new data from various domains over time, and its performance may decline if the data is from a domain that differs significantly from the one it was trained on.

To address this issue, two main approaches have been developed: domain generalization and domain adaptation. Domain generalization~\cite{wang2022generalizing} is a method to enhance a model’s ability to generalize to unseen domains by training the model on labeled data from one or more domains, without assuming any prior knowledge of the test environment in which the model will be applied. However, collecting a large volume of labeled data from various domains for a particular task can be challenging in practice. Single-source domain generalization techniques~\cite{qiao2020learning} have been developed to address this issue, which rely solely on data from a single domain. Although more practical, these techniques generally have lower generalization abilities compared to multi-source domain generalization techniques.

On the other hand, domain adaptation~\cite{wang2018deep} aims to enhance model performance only on the current target domain and does not prioritize performance on all other domains. In particular, unsupervised domain adaptation (UDA)~\cite{patel2015visual} techniques leverage unlabeled data to adapt the model to a new target domain. However, domain adaptation methods fundamentally suffer from performance degradation on a new target domain before and during the adaptation process due to the lack of inherent mechanism to prepare for unseen domains.

\begin{figure*}[!ht]
\begin{center}
\includegraphics[width=\linewidth]{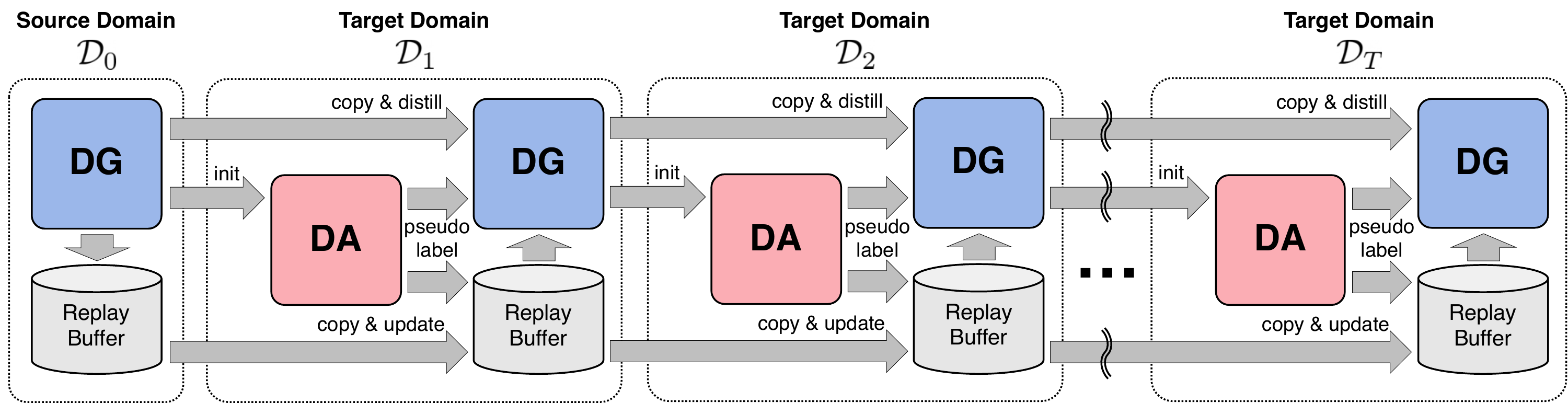}

\end{center}
   \caption{An implementation of CoDAG for unsupervised continual domain shift learning.}
\label{fig:long}

\end{figure*}

In this paper, we tackle a challenging problem that simulates real-world scenarios where models face continual domain shifts and no labeled data is available for new domains. We refer to this problem as \textit{unsupervised continual domain shift learning}. In this setting, the model must continually adapt to new domains (domain adaptation), while maintaining its generalization ability for upcoming and unseen domains (domain generalization), in an unsupervised manner. However, achieving both objectives simultaneously is not always feasible since they involve related but distinct goals. For instance, if the current target domain is vastly dissimilar from any other domains, none of the optimal solutions for adapting to the current target domain with DA would necessarily result in optimal generalization for performing well on other domains. Similarly, achieving optimal generalization for unseen domains through DG may not result in the best solution for the current target domain. Therefore, to address unsupervised continual domain shift learning, it is necessary to find a solution that resolves this trade-off between domain adaptation and generalization.

To address the trade-off between domain adaptation and generalization, we propose \textit{Complementary Domain Adaptation and Generalization (CoDAG)}, a learning framework that combines domain adaptation and domain generalization in a complementary manner. As shown in Fig.~\ref{fig:short}, our approach involves training two separate models: one for domain adaptation and the other for domain generalization. We use the domain adaptation model to adapt to the target domain, generating more accurate and reliable pseudo-labels for training the domain generalization model. In turn, the domain generalization model learns more generalized representations across multiple domains and provides the domain adaptation model with initializing parameters, enhancing its adaptability to a new domain. As a result, the domain adaptation and generalization models complement each other in our framework, leading to improved performance for both. 

The main contribution of our framework, CoDAG, lies in \textit{the complementary manner in which we leverage existing domain adaptation and domain generalization methods to address unsupervised continual domain shift learning}, a unique and challenging problem that has not been thoroughly explored. We deliberately apply existing methods to our framework, rather than introducing new ones, to underscore that the effectiveness of our framework is due to its complementary structure, not its individual components. Indeed, without requiring any models tailored for the present problem, our framework proved its merit by achieving SoTA performance against all baselines, including the one which is explicitly designed for this setting~\cite{liu2023deja}.

Finally, it is important to note that that our work is one of the first attempts to explore the potential synergies between domain adaptation and domain generalization methods. We are breaking new ground by bridging the divide between the disparate fields of domain adaptation and generalization, which were primarily studied independently. This paradigm shift represents not just a novel approach, but one with profound practical implications.

Our contributions can be summarized as follows:
\begin{itemize}

\item We introduce a novel framework that combines domain adaptation and generalization models in a complementary manner, resulting in a synergistic process that enhances overall performance.

\item Our method consistently outperforms state-of-the-art models across all datasets and metrics, demonstrating superior robustness with the lowest standard deviation across different orders in almost all cases.

\item Our approach does not necessitate the use of models designed for the present problem, allowing seamless integration with existing domain adaptation and generalization algorithms for broader applications.
\end{itemize}

\section{Related Work}

\paragraph{Domain generalization}
Domain generalization (DG) is the process of training a model using labeled data from one or multiple domains, with the objective of achieving good generalization performance across unseen domains. Existing DG methods are based on domain-invariant learning~\cite{gan2016learning, ghifary2015domain,  ghifary2016scatter, li2018domain, li2018deep, motiian2017unified}, meta-learning~\cite{balaji2018metareg, dou2019domain, li2018learning, li2019episodic}, and data augmentation~\cite{shankar2018generalizing, zhou2020deep}. To address practical scenarios, single-source DG methods~\cite{qiao2020learning, romera2018train, volpi2018generalizing, zhang2020generalizing, zhao2020maximum} have been proposed, which use labeled data collected from a single domain.  However, these techniques require a large amount of labeled data and can suffer from severe catastrophic forgetting when applied to scenarios with continual domain shift.

\paragraph{Unsupervised domain adaptation}
Unsupervised domain adaptation (UDA)~\cite{patel2015visual} aims to improve the performance of a target model in scenarios where there is a domain shift between the labeled source domain and the unlabeled target domain. UDA methods often achieve distribution alignment through domain invariant feature transformation~\cite{liang2018aggregating, long2013transfer, pan2010domain} or feature space alignment~\cite{fernando2013unsupervised, gopalan2013unsupervised, sun2016return}. Typical domain adaptation techniques generally assume having access to the source data throughout the adaptation process, which is impractical in real-world scenarios. However, source-free domain adaptation (SFDA) methods~\cite{liang2020we,liang2021source} adapt a model to the unlabeled target domain even when the source dataset is not available during the target adaptation process. Recent works on SFDA~\cite{ahmed2021adaptive, chen2022self, li2020model, liu2021source, yeh2021sofa} have arisen, which use generative models to model the distribution of target data by creating pseudo-label refinement~\cite{ahmed2021adaptive, chen2022self}, target-style images~\cite{li2020model, liu2021source}, or variational inference for generating latent source features~\cite{yeh2021sofa}. Nevertheless, these techniques are not intended to accumulate the knowledge acquired from continually shifting domains. 

\paragraph{Continual domain adaptation}
Continual learning (CL)~\cite{de2021continual} focuses on avoiding catastrophic forgetting when learning new tasks by using regularization-based methods~\cite{silver2002task, zenke2017continual} and replay-based methods~\cite{rebuffi2017icarl}. Recent works~\cite{bobu2018adapting, saporta2022multi, tang2021gradient} have adopted the ideas from continual learning to tackle the continual domain adaptation (CDA) problem. These methods involve distilling probability distributions at multiple levels from the previous models to solve the catastrophic forgetting problem~\cite{saporta2022multi}, using sample replay buffers along with domain adversarial training~\cite{bobu2018adapting}, and utilizing a domain-specific memory buffer for each domain~\cite{tang2021gradient}. Despite the use of CDA methods, the model's performance on the target domain is often poor before and during the adaptation process, although it may improve after sufficient adaptation on the new target domain. This can be particularly problematic when there is a significant domain shift. Our complementary framework overcomes this issue by leveraging the DG model's generalization ability to initialize the DA model.

\section{Methodology}
\subsection{Unsupervised continual domain shift learning}
We adopt the problem setting introduced by~\cite{liu2023deja} to address the challenge of \textit{unsupervised continual domain shift learning}. 
Specifically, our approach works with $T+1$ distinct domains $\mathcal{D}_t$ over $t=0,1,\dots,T$ for a given target task, which are sequentially encountered. Here, we tackle the $K$-way image classification problem as the target task, and the image space $\mathcal{X}$ and the label space $\mathcal{Y}$ are shared across all domains. For the sake of notational simplicity, here we use the notation $\mathcal{D}_t$ to denote both the $t$-th domain and the dataset sampled from it interchangeably.

In this setting, the first domain $\mathcal{D}_0$ is regarded as the source domain $\mathcal{S}$ that is used to initially learn the target task. 
Different from other domains, it consists of labeled samples $\mathcal{S}=\mathcal{D}_0 = \{$${(x_0^{(i)}, y_0^{(i)})}$$\}_{i=1}^{N_0}$, where $x_0^{(i)}$ and $y_0^{(i)}$ denote input data and its corresponding label of the $i$-th sample and $N_0$ indicates the total number of samples in the source domain. 
After a model is initially trained with the source domain, it sequentially encounters a series of target domains $\mathcal{T} = \{ \mathcal{D}_1, \mathcal{D}_2, \dots, \mathcal{D}_T \}$. 
In contrast to the source domain, \textit{the target domains do not provide any label information to the model}, as we assume that the model encounters them after the deployment.
Moreover, we consider more realistic settings that \textit{samples from previously experienced domains are not fully accessible} but are partially available via a replay buffer with a limited capacity.
Therefore, the model has to properly improve its generalization ability in an unsupervised manner by mainly using unlabeled samples $\mathcal{D}_t= \{x_t^{(i)}\}_{i=1}^{N_t}$ from the target domain $\mathcal{D}_t$ at each stage $t$.

To deal with this problem setting, a model $f(x;\theta)$ is used to map an input $x \in \mathcal{X}$ to its corresponding label $y \in \mathcal{Y}$, where $\theta$ represents the set of model parameters to be learned. The model is expected to be initially trained with samples from the source domain $\mathcal{S}$ and then adapted to the target domains $\mathcal{T}$ in a continual manner (\ie, from $\mathcal{D}_1$ to $\mathcal{D}_T$). We denote the model parameters after training on the $t$-th domain as $\theta_t^*$ and evaluate the corresponding model $f(x; \theta_t^*)$ with $x$ from different domains. There exist three main goals for this problem setting.
(i) Firstly, we seek to \textit{achieve domain adaptation} for the target domain $\mathcal{D}_t \in \mathcal{T}$. 
(ii) Secondly, we aim to \textit{achieve domain generalization} for unseen target domains $\{\mathcal{D}_{t’}\}_{t<t’\le T}$. 
(iii) Finally, we aim to \textit{prevent catastrophic forgetting} of knowledge gained from previously seen domains $\{\mathcal{D}_{t’}\}_{0\le t’ < t}$.

Most of the existing models for domain adaptation or generalization, which are able to handle unlabeled data, can be applied for unsupervised continual domain shift learning. A naive approach is to start with the model $f(x;\theta_{t-1}^*)$ trained on domain $\mathcal{D}_{t-1}$ and update it with samples from $\mathcal{D}_{t}$, resulting in a new model $f(x; \theta_{t}^*)$ that is either adapted to $\mathcal{D}_{t}$ or generalized for unseen domains $\mathcal{D}_{t’}$ for $t’>t$, depending on its designed purpose. All the baselines used in our experiments reflect this naive approach.

However, achieving both DA and DG objectives in unsupervised continual domain shifts is challenging with a naive single-model approach, given the potential contradiction between adapting to a target domain (DA) and learning domain-invariant features (DG). For instance, when the current target domain is significantly different from any other domains, optimal solutions for domain adaptation may not lead to optimal generalization for other domains. Similarly, using domain generalization to attain optimal generalization for unseen domains does not necessarily yield the best solution for the current target domain.

\subsection{CoDAG: Complementary Domain Adaptation and Generalization }

To resolve this trade-off between DA and DG, we propose a novel framework called \textit{Complementary Domain Adaptation and Generalization (CoDAG)}. CoDAG is highly effective yet remarkably simple. Unlike other approaches that require complicated architectures or learning methods, our CoDAG framework relies on a straightforward and intuitive idea: maintaining two separate models for DA and DG. By having two distinct models optimized for their respective goals, the CoDAG framework eliminates the need for complicated techniques that attempt to balance DA and DG within a single model. Instead, our framework facilitates a synergistic relationship between the two models, where they complement each other, leading to improved performance for both.

To carry out this framework, DA and DG are conducted in an interleaved order with two separate models for DA and DG, and we make each of them to be dependent on the other, as shown in Fig.~\ref{fig:long}. Specifically, we duplicate the model $f$ to use one for DA, denoted as $f_{\text{DA}}$, and the other for DG, denoted as $f_{\text{DG}}$, with each model trained with DA and DG algorithms, respectively.
This setting allows the models to effectively exchange knowledge and continually improve the performance for both.

As domain shifts occur continuously and there is no labeled information available from target domains, nor access to data from previously encountered domains, we have implemented a source-free unsupervised domain adaptation algorithm for our DA model. There are several options for this, but we choose a simple pseudo-labeling~\cite{lee2013pseudo} along with regularization from SHOT~\cite{liang2020we} due to its simplicity. For the DG model, we simply apply an empirical risk minimization (ERM) method along with a RandMix augmentation technique~\cite{liu2023deja} that generates diversified data, which boost the DG model's ability to generalize to unseen domains. However, note that our proposed framework is model-agnostic, in the sense that it can be applied to any DA or DG algorithms that are suitable for the given problem setting. The implementation details of auxiliary methods and algorithms employed in this paper are provided in the Supplementary Material.

\paragraph{Source domain training} 
To initially learn a given target task, we train the DG model from scratch with the set of labeled samples from the source domain $\mathcal{S}$ ($=$ $\mathcal{D}_0$).
This stage can be approached by a \textit{single-source} domain generalization method. In that way, we simply minimize the following cross-entropy loss over the source domain data with enhanced data augmentation,
\begin{equation}
\label{eq:dg}
\mathcal{L}_{\text{DG},0}(\mathcal{D}_0) = \mathbb{E}_{(x,y)\in \mathcal{D}_0} \left[ \mathcal{L}^{\text{ce}}(f_{\text{DG}}(R(x);\theta_{\text{DG},0}), y)\right],
\end{equation}
\noindent where $\theta_{\text{DG},0}$ is the DG model parameters, $\mathcal{L}^{\text{ce}}$ is the cross-entropy loss for a classification setting, and $R$ represents the RandMix augmentation~\cite{liu2023deja}. 

\paragraph{Generalized initialization with DG for DA}
In general, unsupervised source-free domain adaptation approaches involve initially training a source model with the source domain data and further updating the source model with unlabeled data from a new target domain. 
However, in our proposed framework, the DA model utilizes the parameters of the previous DG model for its initialization.
This allows the DA model to leverage the DG model’s generalization ability to learn domain-invariant features and reduce domain-specific factors.
As a result, we achieve efficient adaptation to a new target domain, even when there is a large gap between previously experienced domains and the new target domain (see Section~\ref{sec:init} for experimental results).

To apply this approach for the current target domain $\mathcal{D}_t$, we first initialize the DA model $f_\text{DA}$ with the parameters of the DG model trained with previously experienced domains, or $\theta_{\text{DG},t-1}^*$, treating it as a source model. Then, we freeze the classifier head in $f_\text{DA}$ and only update the feature extractor part of it using information maximization and self-supervised pseudo-labeling with data from the current target domain $\mathcal{D}_t$. Accordingly, the loss to adapt to $\mathcal{D}_t$ is written as,
\begin{equation}
\label{eq:da}
\mathcal{L}_{\text{DA},t}(\mathcal{D}_t)= \mathbb{E}_{x\in \mathcal{D}_t} \left[ \mathcal{L}^{\text{shot}}(f_\text{DA}(x;\theta_{\text{DA},t})) \right],
\end{equation}
\noindent where $\theta_{\text{DA},t}$ is the parameters of the DA model initialized with the optimal parameters of the DG model trained on the previous domain $\mathcal{D}_{t-1}$, or $\theta_{\text{DG},t-1}^{*}$, and $\mathcal{L}^{\text{shot}}$ is the cross-entropy loss with pseudo-labels on target predictions along with regularization from SHOT~\cite{liang2020we}.

\paragraph{Pseudo-label generation with DA for DG}
In our proposed framework, we simply use an empirical risk minimization (ERM) method along with an enhanced data augmentation method for training the DG model. Although labeled samples are necessary for this process, none of the target domains $\mathcal{T}$ provide any labels. Thus, to make use of unlabeled samples from the current target domain $\mathcal{D}_t$, we adopt a pseudo-label generation strategy~\cite{lee2013pseudo} based on the highest prediction confidence of the DA model adapted to $\mathcal{D}_t$. This involves applying the DA model to the unlabeled samples from $\mathcal{D}_t$ to generate pseudo-labels, which are then used as training labels for the DG model on $\mathcal{D}_t$.

Specifically, for each unlabeled sample $x$, we compute its pseudo-label $\hat{y}_t(x)$ as follows:
\begin{equation}
  \hat{y}_{t}(x) =  \argmax_{k}\delta_{k}(f_\text{DA}(x;\theta_{\text{DA},t}^*)),
\end{equation}
where $\theta_{\text{DA},t}^*$ is obtained by optimizing $\mathcal{L}_{\text{DA},t}$ in Eq.~\ref{eq:da} and $\delta_k(\cdot)$ is the $k$-th element of a softmax output.
By using the resulting pseudo-labels as training labels for the DG model on $\mathcal{D}_t$, we construct a pseudo-labeled dataset $\hat{\mathcal{D}}_t=\{(x_t^{(i)},\hat{y}_t(x_t^{(i)}))$$\}_{i=1}^{N_t}$. We then update the DG model with ERM in the same way as the source training in Eq.~\ref{eq:dg} with the following loss:
\begin{equation}
\label{eq:pl}
\mathcal{L}_{\text{DG},t}^{\text{erm}}(\hat{\mathcal{D}}_t) = \mathbb{E}_{(x,y) \in \hat{\mathcal{D}}_t} \left[\mathcal{L}^{\text{ce}}(f_\text{DG}(R(x);\theta_{\text{DG},t}),y)\right],
\end{equation}
\noindent where $\theta_{\text{DG},t}$ is initialized with the optimal parameters of the DG model trained on the previous domain $\mathcal{D}_{t-1}$, or $\theta_{\text{DG},t-1}^{*}$.

\begin{table*}[!ht]
\caption{Comparison of the performance on the PACS, Digits-five, and DomainNet datasets for different state-of-art methods in TDA, TDG, FA, and All. The results are averaged over 10 different orders from each dataset. The results of the baseline models are referenced from~\cite{liu2023deja}. The best results are highlighted in bold. Our method outperforms all the baselines across all datasets and evaluation metrics tested.}
\label{tab:1}
\begin{center}

\begin{tabular}{c|c|cccccccc|c}

\hline
\multirow{3}{*}{Dataset} & \multirow{3}{*}{Metric} & \multicolumn{8}{c|}{Comparison Baselines (\textit{w/ Replay Buffer})} & Ours \\ \cline{3-11} 
 &  & SHOT+ & SHOT++ & Tent & AdaCon & EATA & L2D & PDEN & RaTP & \multirow{2}{*}{CoDAG} \\
 &  & ~\cite{liang2021source,liu2023deja} & ~\cite{liang2021source} & ~\cite{wang2020tent} & ~\cite{chen2022contrastive} & ~\cite{niu2022efficient} & ~\cite{wang2021learning} & ~\cite{li2021progressive} & ~\cite{liu2023deja} &  \\ \hline \hline
\multirow{8}{*}{PACS} 
 & \multirow{2}{*}{TDA} & 81.9 & 84.4 & 78.7 & 79.9 & 80.3 & 78.8 & 77.8 & 84.7 & \textbf{87.6} \\
 &  & $\scalemath{0.8}{\pm}\text{\footnotesize9.2}$ & $\scalemath{0.8}{\pm}\text{\footnotesize8.0}$ & $\scalemath{0.8}{\pm}\text{\footnotesize6.9}$ & $\scalemath{0.8}{\pm}\text{\footnotesize5.9}$ & $\scalemath{0.8}{\pm}\text{\footnotesize7.1}$ & $\scalemath{0.8}{\pm}\text{\footnotesize5.6}$ & $\scalemath{0.8}{\pm}\text{\footnotesize5.2}$ & $\scalemath{0.8}{\pm}\text{\footnotesize5.1}$ & $\scalemath{0.8}{\pm}\text{\footnotesize4.0}$ \\
 & \multirow{2}{*}{TDG} & 54.9 & 56.0 & 65.8 & 65.2 & 64.1 & 65.8 & 64.4 & 70.6 & \textbf{72.2} \\
 &  & $\scalemath{0.8}{\pm}\text{\footnotesize13.1}$ & $\scalemath{0.8}{\pm}\text{\footnotesize10.9}$ & $\scalemath{0.8}{\pm}\text{\footnotesize11.5}$ & $\scalemath{0.8}{\pm}\text{\footnotesize10.5}$ & $\scalemath{0.8}{\pm}\text{\footnotesize12.1}$ & $\scalemath{0.8}{\pm}\text{\footnotesize9.6}$ & $\scalemath{0.8}{\pm}\text{\footnotesize9.8}$ & $\scalemath{0.8}{\pm}\text{\footnotesize9.1}$ & $\scalemath{0.8}{\pm}\text{\footnotesize8.3}$ \\
 & \multirow{2}{*}{FA} & 74.9 & 83.0 & 81.0 & 81.6 & 82.6 & 77.6 & 76.3 & 83.9 & \textbf{88.8} \\
 &  & $\scalemath{0.8}{\pm}\text{\footnotesize8.1}$ & $\scalemath{0.8}{\pm}\text{\footnotesize4.0}$ & $\scalemath{0.8}{\pm}\text{\footnotesize6.2}$ & $\scalemath{0.8}{\pm}\text{\footnotesize5.9}$ & $\scalemath{0.8}{\pm}\text{\footnotesize7.0}$ & $\scalemath{0.8}{\pm}\text{\footnotesize4.6}$ & $\scalemath{0.8}{\pm}\text{\footnotesize4.0}$ & $\scalemath{0.8}{\pm}\text{\footnotesize4.7}$ & $\scalemath{0.8}{\pm}\text{\footnotesize3.0}$ \\ \cline{2-11} 
 & \multirow{2}{*}{All} & 70.6 & 74.5 & 75.2 & 75.6 & 75.7 & 74.1 & 72.9 & 79.7 & \textbf{82.9} \\
 &  & $\scalemath{0.8}{\pm}\text{\footnotesize9.2}$ & $\scalemath{0.8}{\pm}\text{\footnotesize5.7}$ & $\scalemath{0.8}{\pm}\text{\footnotesize7.8}$ & $\scalemath{0.8}{\pm}\text{\footnotesize7.1}$ & $\scalemath{0.8}{\pm}\text{\footnotesize8.6}$ & $\scalemath{0.8}{\pm}\text{\footnotesize6.2}$ & $\scalemath{0.8}{\pm}\text{\footnotesize5.9}$ & $\scalemath{0.8}{\pm}\text{\footnotesize5.7}$ & $\scalemath{0.8}{\pm}\text{\footnotesize4.8}$ \\ \hline \hline
\multirow{8}{*}{Digits-five} 
 & \multirow{2}{*}{TDA} & 78.6 & 81.3 & 68.7 & 71.6 & 72.0 & 84.3 & 82.3 & 88.7 & \textbf{92.7} \\
 &  & $\scalemath{0.8}{\pm}\text{\footnotesize13.2}$ & $\scalemath{0.8}{\pm}\text{\footnotesize14.0}$ & $\scalemath{0.8}{\pm}\text{\footnotesize11.0}$ & $\scalemath{0.8}{\pm}\text{\footnotesize9.2}$ & $\scalemath{0.8}{\pm}\text{\footnotesize9.8}$ & $\scalemath{0.8}{\pm}\text{\footnotesize5.4}$ & $\scalemath{0.8}{\pm}\text{\footnotesize5.8}$ & $\scalemath{0.8}{\pm}\text{\footnotesize1.8}$ & $\scalemath{0.8}{\pm}\text{\footnotesize1.7}$ \\
 & \multirow{2}{*}{TDG} & 61.0 & 62.3 & 64.0 & 63.3 & 64.0 & 70.9 & 69.7 & 76.8 & \textbf{77.4} \\
 &  & $\scalemath{0.8}{\pm}\text{\footnotesize14.9}$ & $\scalemath{0.8}{\pm}\text{\footnotesize13.8}$ & $\scalemath{0.8}{\pm}\text{\footnotesize13.6}$ & $\scalemath{0.8}{\pm}\text{\footnotesize13.1}$ & $\scalemath{0.8}{\pm}\text{\footnotesize12.9}$ & $\scalemath{0.8}{\pm}\text{\footnotesize6.8}$ & $\scalemath{0.8}{\pm}\text{\footnotesize7.0}$ & $\scalemath{0.8}{\pm}\text{\footnotesize3.9}$ & $\scalemath{0.8}{\pm}\text{\footnotesize4.3}$ \\
 & \multirow{2}{*}{FA} & 58.2 & 64.5 & 66.1 & 72.2 & 73.0 & 76.5 & 74.0 & 85.0 & \textbf{87.1} \\
 &  & $\scalemath{0.8}{\pm}\text{\footnotesize14.9}$ & $\scalemath{0.8}{\pm}\text{\footnotesize13.3}$ & $\scalemath{0.8}{\pm}\text{\footnotesize15.7}$ & $\scalemath{0.8}{\pm}\text{\footnotesize11.2}$ & $\scalemath{0.8}{\pm}\text{\footnotesize10.9}$ & $\scalemath{0.8}{\pm}\text{\footnotesize3.8}$ & $\scalemath{0.8}{\pm}\text{\footnotesize4.0}$ & $\scalemath{0.8}{\pm}\text{\footnotesize2.2}$ & $\scalemath{0.8}{\pm}\text{\footnotesize2.1}$ \\ \cline{2-11} 
 & \multirow{2}{*}{All} & 65.9 & 69.4 & 66.2 & 69.1 & 69.6 & 77.2 & 75.3 & 83.5 & \textbf{85.7} \\
 &  & $\scalemath{0.8}{\pm}\text{\footnotesize13.5}$ & $\scalemath{0.8}{\pm}\text{\footnotesize12.9}$ & $\scalemath{0.8}{\pm}\text{\footnotesize13.3}$ & $\scalemath{0.8}{\pm}\text{\footnotesize11.0}$ & $\scalemath{0.8}{\pm}\text{\footnotesize10.9}$ & $\scalemath{0.8}{\pm}\text{\footnotesize4.8}$ & $\scalemath{0.8}{\pm}\text{\footnotesize5.1}$ & $\scalemath{0.8}{\pm}\text{\footnotesize2.1}$ & $\scalemath{0.8}{\pm}\text{\footnotesize2.2}$ \\ \hline \hline
\multirow{8}{*}{\begin{tabular}[c]{@{}c@{}}DomainNet\end{tabular}} 
 & \multirow{2}{*}{TDA} & 66.0 & 66.9 & 53.6 & 62.2 & 62.5 & 56.2 & 55.6 & 65.4 & \textbf{71.0} \\
 &  & $\scalemath{0.8}{\pm}\text{\footnotesize8.8}$ & $\scalemath{0.8}{\pm}\text{\footnotesize8.7}$ & $\scalemath{0.8}{\pm}\text{\footnotesize13.2}$ & $\scalemath{0.8}{\pm}\text{\footnotesize7.7}$ & $\scalemath{0.8}{\pm}\text{\footnotesize7.3}$ & $\scalemath{0.8}{\pm}\text{\footnotesize6.2}$ & $\scalemath{0.8}{\pm}\text{\footnotesize6.6}$ & $\scalemath{0.8}{\pm}\text{\footnotesize5.1}$ & $\scalemath{0.8}{\pm}\text{\footnotesize5.7}$ \\
 & \multirow{2}{*}{TDG} & 47.3 & 48.1 & 47.7 & 51.3 & 52.1 & 50.7 & 49.3 & 55.2 & \textbf{56.2} \\
 &  & $\scalemath{0.8}{\pm}\text{\footnotesize11.0}$ & $\scalemath{0.8}{\pm}\text{\footnotesize10.7}$ & $\scalemath{0.8}{\pm}\text{\footnotesize11.0}$ & $\scalemath{0.8}{\pm}\text{\footnotesize10.0}$ & $\scalemath{0.8}{\pm}\text{\footnotesize9.9}$ & $\scalemath{0.8}{\pm}\text{\footnotesize9.1}$ & $\scalemath{0.8}{\pm}\text{\footnotesize9.1}$ & $\scalemath{0.8}{\pm}\text{\footnotesize7.4}$ & $\scalemath{0.8}{\pm}\text{\footnotesize7.2}$ \\
 & \multirow{2}{*}{FA} & 58.5 & 66.9 & 56.1 & 61.8 & 62.8 & 52.2 & 50.2 & 63.5 & \textbf{70.9} \\
 &  & $\scalemath{0.8}{\pm}\text{\footnotesize8.3}$ & $\scalemath{0.8}{\pm}\text{\footnotesize6.0}$ & $\scalemath{0.8}{\pm}\text{\footnotesize14.5}$ & $\scalemath{0.8}{\pm}\text{\footnotesize9.0}$ & $\scalemath{0.8}{\pm}\text{\footnotesize8.8}$ & $\scalemath{0.8}{\pm}\text{\footnotesize9.4}$ & $\scalemath{0.8}{\pm}\text{\footnotesize9.5}$ & $\scalemath{0.8}{\pm}\text{\footnotesize6.6}$ & $\scalemath{0.8}{\pm}\text{\footnotesize6.6}$ \\ \cline{2-11} 
 & \multirow{2}{*}{All} & 57.3 & 60.6 & 52.5 & 58.4 & 59.1 & 53.0 & 51.7 & 61.4 & \textbf{66.0} \\
 &  & $\scalemath{0.8}{\pm}\text{\footnotesize8.9}$ & $\scalemath{0.8}{\pm}\text{\footnotesize8.0}$ & $\scalemath{0.8}{\pm}\text{\footnotesize12.4}$ & $\scalemath{0.8}{\pm}\text{\footnotesize8.6}$ & $\scalemath{0.8}{\pm}\text{\footnotesize8.3}$ & $\scalemath{0.8}{\pm}\text{\footnotesize7.6}$ & $\scalemath{0.8}{\pm}\text{\footnotesize7.8}$ & $\scalemath{0.8}{\pm}\text{\footnotesize6.0}$ & $\scalemath{0.8}{\pm}\text{\footnotesize6.2}$ \\ 
 \hline

\end{tabular}
\end{center} 
\end{table*}

\paragraph{Learning from noisy labels}
As the DA model is optimized to adapt to the current domain $\mathcal{D}_t$, we assume that it can generate high-quality pseudo-labels. However, some of the labels may still contain errors due to the imperfectness of the DA method. Unfortunately, the errors in pseudo-labels (\ie, noisy labels~\cite{song2022learning}) can negatively affect the performance of the DG model. To alleviate this problem, we use an algorithm that can properly handle noisy labels to prevent the performance degradation of the DG model.

In this paper, we adopt an algorithm called Selective Negative Learning and Positive Learning (SelNLPL)~\cite{kim2019nlnl} to reduce the risk of overfitting to noisy labels and improve performance of the DG model. We show the effectiveness of this approach in Sec.~\ref{sel}. Note that our approach offers greater flexibility that is not restricted to SelNLPL but rather can leverage any label noise-resilient methods~\cite{li2017learning, rizve2021defense, van2015learning}.

\paragraph{Forgetting alleviation}
Forgetting alleviation is crucial in unsupervised continual domain shift learning, where limited access to data from previous domains makes it challenging to maintain performance on previous tasks. When encountering a new target domain, the performance of the DG model on previous domains tends to degrade, which is commonly referred to as catastrophic forgetting in continual learning.

To address this issue, we add a simple distillation loss term~\cite{hinton2015distilling} $\mathcal{L}_{\text{DG},t}^{\text{distill}}$ to the loss of the DG model in Eq.~\ref{eq:pl} to ensure that the model retains the knowledge gained from previous domains $\{\mathcal{D}_{t'}\}_{0 \leq t' < t}$ while learning from the current target domain $\mathcal{D}_t$, given by
\begin{equation}
\mathcal{L}_{\text{DG},t}^{\text{distill}}(\mathcal{D}_t) = \mathbb{E}_{x\in \mathcal{D}_t} \left[\mathcal{L}^{\text{kl}}(q_{t}(x)||p_{t}(x)) \right],
\end{equation}
\noindent where $\mathcal{L}^{\text{kl}}$ represents the KL divergence loss, and $q_{t}(x)=\delta(f_\text{DG}(R(x);\theta_{\text{DG},{t-1}}^*))$ and $p_{t}(x)=\delta(f_{\text{DG}}(R(x);\theta_{\text{DG},t}))$ are the predicted softmax probabilites from the previous and current DG models, respectively.

Another method we employ to prevent catastrophic forgetting is a replay buffer $\mathcal{M}_t$ of size $M \ll N_t$~\cite{rolnick2019experience}, which contains selected samples from previously experienced domains $\{\mathcal{D}_{t'}\}_{0 \leq t' < t}$. We build the replay buffer based on the iCaRL approach~\cite{liu2023deja, rebuffi2017icarl}. By using a replay buffer, along with the samples from $\mathcal{D}_{t}$, additional $M$ selected samples from $\mathcal{D}_0\cup\{\hat{\mathcal{D}}_{t'}\}_{1 \leq t' < t}$ are available for training the DG model on $\mathcal{D}_{t}$. This allows the DG model to not only improve its generalization ability but also prevent catastrophic forgetting. Then, our final loss to update the DG model on the current target domain $\mathcal{D}_{t}$ is given by,
\begin{equation}
\mathcal{L}_{\text{DG},t}(\tilde{\mathcal{D}}_{t}) = \mathcal{L}_{\text{DG},t}^{\text{erm}}(\tilde{\mathcal{D}}_{t}) + \alpha \cdot \mathcal{L}_{\text{DG},t}^{\text{distill}}(\tilde{\mathcal{D}}_{t}),
\end{equation}
\noindent where $\tilde{\mathcal{D}}_{t}=\hat{\mathcal{D}}_{t}\cup \mathcal{M}_t$ represents $N_t+M$ samples available from the $t$-th domain with a replay buffer and $\alpha$ is a balancing hyperparameter.

Our findings suggest that utilizing a replay buffer leads to a significant improvement in performance, especially when dealing with previous domains. However, even without a replay buffer, our experiment shows that our model remains competitive against state-of-the-art models that are equipped with replay buffers (see Sec.~\ref{ab} for experimental results).

\section{Experiments}

To validate the effectiveness of our framework, we compare our proposed framework, CoDAG, against state-of-the-art methods on three benchmark datasets: (i) PACS~\cite{li2017deeper}, (ii) Digits-five~\cite{ganin2015unsupervised,hull1994database,lecun1998gradient, netzer2011reading}, and (iii) DomainNet~\cite{peng2019moment}. For a fair comparison, we adhere to the experimental setup from~\cite{liu2023deja} in the following sections.

\paragraph{Datasets}
PACS consists of 4 distinct domains with 7 classes, including Photo (P), Art painting (A), Cartoon (C), and Sketch (S). Digits-five contains 5 different domains with 10 classes, 0 to 9, including MNIST (MT)~\cite{lecun1998gradient}, MNIST-M (MM)~\cite{ganin2015unsupervised}, SVHN (SN)~\cite{netzer2011reading}, SYN-D (SD)~\cite{ganin2015unsupervised} and USPS (US)~\cite{hull1994database}. DomainNet is the most challenging dataset, which includes Quickdraw (Qu), Clipart (Cl), Painting (Pa), Infograph (In), Sketch (Sk) and Real (Re). DomainNet has an imbalance in class distribution, where some domains have limited images for certain classes. To address this issue, a subset of DomainNet is used by selecting the top 10 classes with most images in the whole dataset.

\paragraph{Experimental settings}
In the source domain, 80\% of the data is randomly assigned as a training set, and the remaining 20\% as a testing set. In target domains, all data is used for training and testing as we assume any label information is not available. The experiments are repeated three times using different seeds (2022, 2023, 2024), and the average performance is reported. For Digits-five, we use DTN~\cite{liang2020we} as a feature extractor, while for both PACS and DomainNet, we employ ResNet-50~\cite{he2016deep}. The SGD optimizer is used with a batch size of 64 for all experiments. The size of replay buffer is set to 200 for all datasets. See the Supplementary Material for more details on the experimental settings and network architectures.

\begin{figure}[t]
\begin{center}
\includegraphics[width=\linewidth]{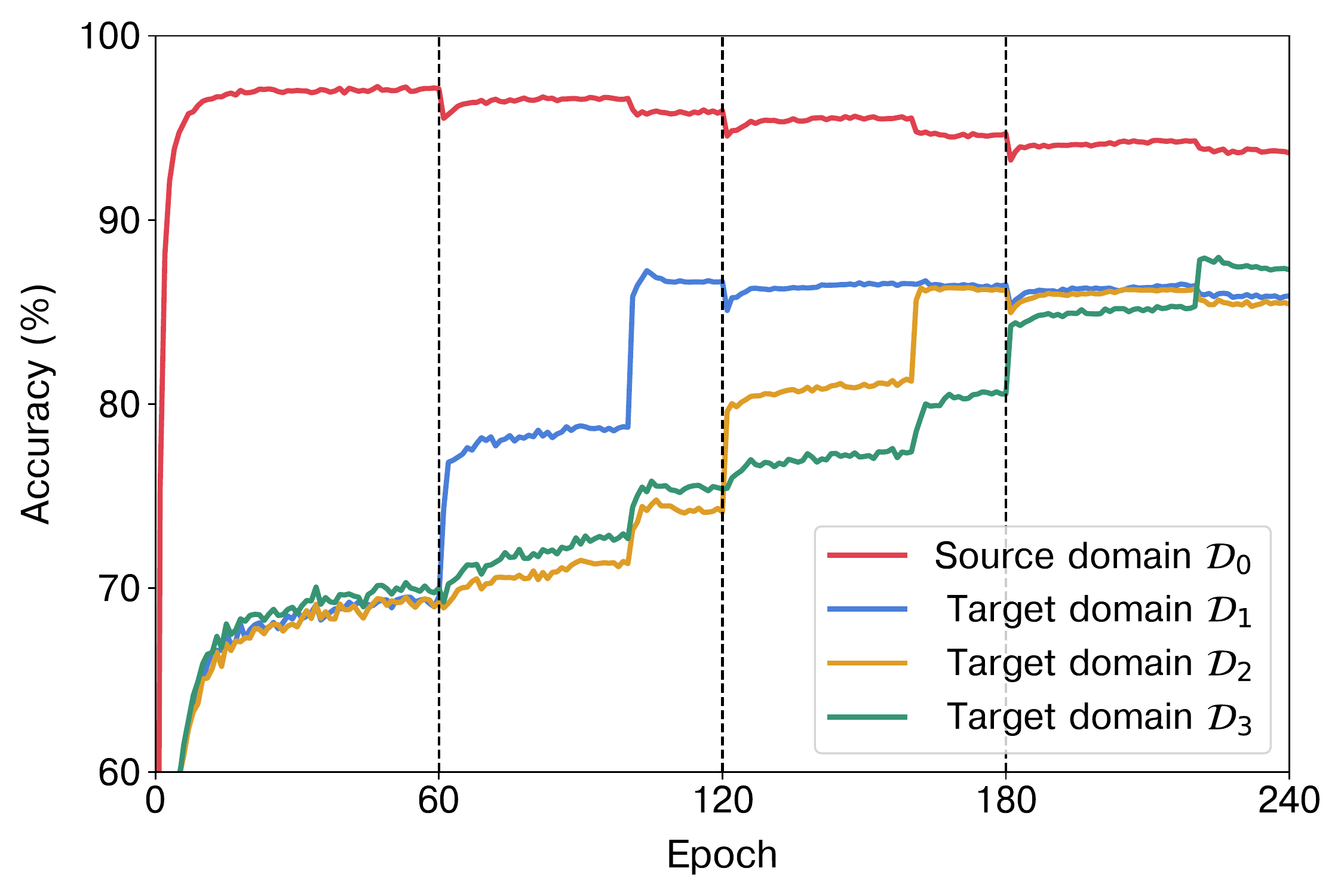}
\caption{The training curves of the DG model which depict the model's performance on their respective domains at every training epoch. The domain shift occurs every 60 epochs, starting from the source domain and continuing until the third target domain. The results are averaged over 10 different orders from the PACS dataset.}
\label{fig:train}
\end{center}
\end{figure}

\paragraph{Evaluation metrics}
(i) TDA: \textit{Target domain adaptation} for each domain is the performance measured on the domain right after its training stage. The TDA of the $t$-th domain for $t=0,\dots,T$ is 
\begin{equation}
\text{TDA}_t = \mathcal{A}(f(x;\theta_{t}^*),\mathcal{D}_t),
\end{equation}
\noindent where $\mathcal{A}(f(x;\theta_{t'}^*),\mathcal{D}_t)$ represents the test accuracy on the domain $\mathcal{D}_t$ with the model $f(x;\theta_{t'}^*)$ obtained after training on the domain $\mathcal{D}_{t'}$.
(ii) TDG: \textit{Target domain generalization} for each domain is evaluated by the average performance on the domain prior to its training stage. The TDG of the $t$-th domain for $t=1,\dots,T$ is given by
\begin{equation}
\text{TDG}_t = \frac{1}{t}\sum_{t'=0}^{t-1}\mathcal{A}(f(x;\theta_{t'}^*),\mathcal{D}_t).
\end{equation}
\noindent (iii) FA: \textit{Forgetting alleviation} for each domain is evaluated by the average performance on the domain after the model has been trained on subsequent domains. The FA of the $t$-th domain for $t=0,\dots,T-1$ can be written as
\begin{equation}
\text{FA}_t = \frac{1}{T-t}\sum_{t'=t+1}^{T}\mathcal{A}(f(x;\theta_{t'}^*),\mathcal{D}_t).
\end{equation}
For comparative analysis between different models, we use the averaged value of each metric over all domains for which the metric is defined. Additionally, we define the average of all the metrics as a composite score (All) to evaluate the overall performance of the models. To properly evaluate our proposed framework, CoDAG, we use the DA model $f_\text{DA}$ to evaluate TDA, while the DG model $f_\text{DG}$ is used to evaluate TDG and FA.

\vspace{1.0em}
\subsection{Effectiveness of the CoDAG framework}

\paragraph{Comparing with state-of-the-art}
To compare the performance of our model with state-of-the-art methods, we use the experiment results of different methods reported in~\cite{liu2023deja} as baselines. These methods include several state-of-the-art models from Source-Free DA (SHOT+~\cite{liang2021source,liu2023deja} and SHOT++~\cite{liang2021source}), Test-Time/Online DA (Tent \cite{wang2020tent}, AdaCon \cite{chen2022contrastive}, and EATA~\cite{niu2022efficient}), Single DG (L2D~\cite{wang2021learning} and PDEN~\cite{li2021progressive}), as well as Continual DG (RaTP~\cite{liu2023deja}). To ensure consistency and fairness, we adopt the same backbone feature extractor with the baseline methods. In the present experiments, all baseline methods are equipped with the replay buffer of size $200$.

We present an evaluation of our CoDAG framework against the baseline models on the three datasets (PACS, Digits-five, and DomainNet) using the four evaluation metrics (TDA, TDG, FA, and All). The results are averaged over 10 different orders from each dataset to ensure the robustness of our findings. The overall results presented in Table~\ref{tab:1} demonstrate that our method consistently outperforms all the baseline models across all datasets and evaluation metrics.

Compared to RaTP~\cite{liu2023deja}, a continual domain generalization method that achieves the best performance among the baselines in most cases, our method demonstrates significantly higher performance in TDA and FA. This highlights the effectiveness of our domain adaptation stage based on the generalized initialization using the DG model. The more accurate pseudo-labels generated by the DA model also contribute to the DG model's superior performance in FA.

On the DomainNet dataset, which is the most challenging among our benchmark datasets, the DA-specialized model SHOT++~\cite{liang2021source} outperforms RaTP in terms of TDA and FA. However, our CoDAG consistently outperforms any DA-specialized models in terms of TDA and FA, while maintaining the improved generalization performance in terms of TDG.

Furthermore, Table~\ref{tab:1} shows that our method acheives the lowest standard deviation across ten different orders in nearly all cases. Notably, even our lower bound performance $(\mu - \sigma)$ surpasses the upper bound performance $(\mu + \sigma)$ of other baselines in many cases. These findings demonstrate the robustness of our CoDAG framework in addressing the challenges of unsupervised continual domain shift learning.

\begin{table}[t]
\begin{center}
\caption{Evaluation of two distinct initialization methods for the DA model with performance averaged across 10 different orders from the PACS dataset.}
\vspace{1.em}
\label{tab:init}
\begin{tabular}{c|ccc|c}
\hline
Method & TDA & TDG & FA & All \\ \hline \hline
\multirow{2}{*}{\begin{tabular}[c]{@{}c@{}}Initialization w/\\ the DG model\end{tabular}} & 87.6 & 72.2 & 88.8 & 82.9 \\
 & $\scalemath{0.8}{\pm}\text{\footnotesize4.0}$ & $\scalemath{0.8}{\pm}\text{\footnotesize8.3}$ & $\scalemath{0.8}{\pm}\text{\footnotesize3.0}$ & $\scalemath{0.8}{\pm}\text{\footnotesize4.8}$ \\ \hline
\multirow{2}{*}{\begin{tabular}[c]{@{}c@{}}Initialization w/\\ the DA model\end{tabular}} & 83.8 & 71.7 & 86.6 & 80.7 \\
 & $\scalemath{0.8}{\pm}\text{\footnotesize4.0}$ & $\scalemath{0.8}{\pm}\text{\footnotesize8.4}$ & $\scalemath{0.8}{\pm}\text{\footnotesize4.0}$ & $\scalemath{0.8}{\pm}\text{\footnotesize4.8}$ \\ \hline
Diff. & +3.8 & +0.5 & +2.2 & +2.2 \\ 
\hline
\end{tabular}
\end{center}
\end{table}

\begin{table*}[t]
\caption{The ablation study of SelNLPL conducted for all possible pairs of two domains ($\mathcal{D}_0 \rightarrow \mathcal{D}_1$) from the PACS dataset. TDG was measured by the average performance on two unseen domains ($\mathcal{D}_2$ and $\mathcal{D}_3$), while FA was measured by the performance on the source domain ($\mathcal{D}_0$). Diff. denotes the result obtained by subtracting the performance without (w/o) SelNLPL from the performance with (w/) SelNLPL.}
\vspace{0.5em}
\label{table:2}
\begin{center}
\footnotesize
\begin{tabular}{c|c|cccccccccccc|c}
\hline

Metric               & Method      & P$\rightarrow$A & P$\rightarrow$C & P$\rightarrow$S & A$\rightarrow$P & A$\rightarrow$C & A$\rightarrow$S & C$\rightarrow$P & C$\rightarrow$A & C$\rightarrow$S & S$\rightarrow$P & S$\rightarrow$A & S$\rightarrow$C & Avg. \\ \hline \hline
\multirow{3}{*}{TDG} & w/ SelNLPL  & 58.3 & 74.6 & 58.4 & 59.8 & 85.3 & 77.7 & 79.8 & 86.8 & 78.1 & 69.9 & 84.3 & 85.3 & 74.9 \\
                     & w/o SelNLPL & 57.4 & 74.2 & 54.3 & 58.0 & 84.9 & 71.2 & 78.6 & 86.6 & 74.2 & 68.3 & 83.4 & 85.0 & 73.0 \\ \cline{2-15} 
                     & Diff.       & +0.9 & +0.4 & +4.1 & +1.8 & +0.4 & +6.5 & +1.2 & +0.2 & +3.9 & +1.6 & +0.9 & +0.3 & +1.8 \\ \hline
\multirow{3}{*}{FA}  & w/ SelNLPL  & 98.5 & 98.2 & 95.7 & 94.1 & 93.2 & 83.0 & 90.4 & 92.6 & 85.8 & 85.7 & 91.2 & 89.7 & 91.5 \\
                     & w/o SelNLPL & 98.5 & 97.2 & 93.0 & 93.2 & 92.6 & 76.7 & 89.4 & 91.3 & 84.7 & 85.1 & 89.3 & 88.4 & 89.9 \\ \cline{2-15} 
                     & Diff.       & 0.0  & +1.0 & +2.7 & +0.9 & +0.6 & +6.3 & +1.0 & +1.3 & +1.1 & +0.6 & +1.9 & +1.3 & +1.6 \\ 
\hline

\end{tabular}
\end{center}
\end{table*}

\paragraph{Training curves}
In Fig.~\ref{fig:train}, we display the training curves of the DG model, each of which represents the accuracy of the model on its corresponding domain at different training stages. We observe that the model's performance on unseen domains gradually increases as training progresses, which illustrates the model's continually improving generalization ability. Moreover, Fig.~\ref{fig:train} suggests that the model is capable of avoiding catastrophic forgetting, as its performance on previously encountered domains remains relatively stable even after domain shifts.

\vspace{0.8em}
\subsection{Further analysis}
\label{further}
In this section, we perform additional analyses to investigate
the key components of our method using the PACS dataset under different experimental settings.

\paragraph{Effectiveness of generalized initialization for DA}
\label{sec:init}
To evaluate the effectiveness of the generalized initialization approach for DA, we compare model performance between two different approaches for initializing $\theta_{\text{DA},t}$ of the DA model on $\mathcal{D}_t$: (1) initializing with the previous DG model using $\theta_{\text{DG},t-1}^*$ and (2) initializing with the previous DA model using $\theta_{\text{DA},t-1}^*$. The results presented in Table~\ref{tab:init} show that initializing with the DG model significantly improves the performance in TDA. This exhibits the effectiveness of the generalized initialization, which enables the DA model to adapt more efficiently to a new domain, compared to relying on parameters specifically adapted to the previous domain. Furthermore, we observe that the improved TDA of the DA model has a positive impact on the performance of the DG model in FA and TDG by providing more accurate pseudo-labels.

\begin{figure}[t]
\begin{center}
\includegraphics[width=\linewidth]{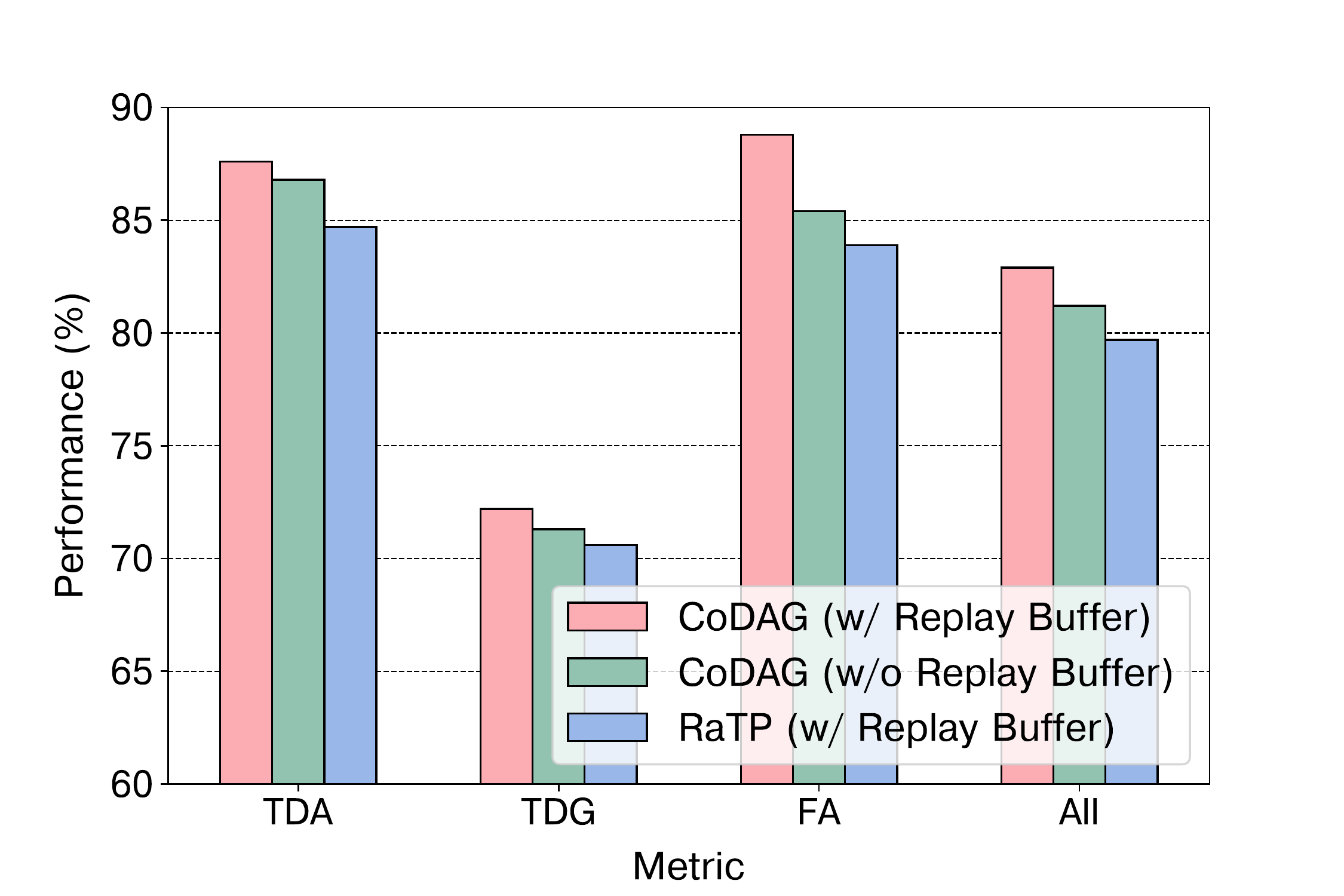}
\end{center}
\caption{The ablation study of replay buffer with averaged performance across 10 different orders from the PACS dataset.}
\label{buffer}
\end{figure}

\paragraph{Ablation study of SelNLPL}
\label{sel}
To understand SelNLPL's contribution, we conduct experiments with and without SelNLPL in the training of our model for pairs of two different domains ($\mathcal{D}_0 \rightarrow \mathcal{D}_1$). We then compare the resulting changes in TDG and FA. To isolate the effect of SelNLPL, we remove the replay buffer. The experiment results in Table~\ref{table:2} demonstrate that SelNLPL can improve model performance in both TDG and FA, which underscores the effectiveness of noise-resilient methods to amplify the complementary effect of the DA model on the DG model within our framework.

\paragraph{Ablation study of replay buffer}
\label{ab}
We perform ablation studies to assess the effectiveness of utilizing a replay buffer in CoDAG, by comparing the performance of the model with and without the buffer. The results presented in Fig.~\ref{buffer} clearly display that the absence of the buffer leads to a degradation in performance across all metrics, with the metric for FA being the most adversely affected. This indicates that the replay buffer plays a crucial role in preventing catastrophic forgetting.

Furthermore, our findings suggest that the replay buffer enhances the model's ability to generalize by continually exposing it to multiple domains, thereby boosting its performance in TDG. This, in turn, leads to an improvement in TDA as the model can adapt to a new domain with more generalized initialization.

However, despite the decrease in performance after the removal of the buffer, our method \textit{without replay buffer} still outperforms all other baselines \textit{with replay buffer}, further confirming the effectiveness of our framework.

\section{Conclusion}
In this paper, we propose a learning framework that combines domain adaptation and generalization models in a complementary manner, which effectively addresses the challenge of unsupervised continual domain shift learning. Our method outperforms state-of-the-art performance across different datasets and evaluation metrics. It also achieves competitive results even without a replay buffer, demonstrating its effectiveness and robustness in real-world scenarios. 

Our approach is model-agnostic, meaning it can be used with any domain adaptation and generalization algorithms suitable for a given problem. We envision extending our method to complex scenarios beyond a pre-defined set of domains, dynamically discovering domain shifts and adapting to new domains. This can increase its applicability to a broader range of scenarios, with potential contributions to practical applications in computer vision and other fields.

\section*{Acknowledgement}
This work was supported by the Hyundai Motor Chung Mong-Koo Foundation, the New Faculty Startup Fund from Seoul National University, and IITP (RS-2023-00232046).

{\small
\bibliographystyle{ieee_fullname}
\bibliography{main}

\begin{thebibliography}{10}\itemsep=-1pt

\bibitem{ahmed2021adaptive}
Waqar Ahmed, Pietro Morerio, and Vittorio Murino.
\newblock Adaptive pseudo-label refinement by negative ensemble learning for source-free unsupervised domain adaptation.
\newblock {\em arXiv preprint arXiv:2103.15973}, 2021.

\bibitem{balaji2018metareg}
Yogesh Balaji, Swami Sankaranarayanan, and Rama Chellappa.
\newblock Metareg: Towards domain generalization using meta-regularization.
\newblock {\em Advances in neural information processing systems}, 31, 2018.

\bibitem{bengio2000neural}
Yoshua Bengio, R{\'e}jean Ducharme, and Pascal Vincent.
\newblock A neural probabilistic language model.
\newblock {\em Advances in neural information processing systems}, 13, 2000.

\bibitem{bobu2018adapting}
Andreea Bobu, Eric Tzeng, Judy Hoffman, and Trevor Darrell.
\newblock Adapting to continuously shifting domains.
\newblock 2018.

\bibitem{bojarski2016end}
Mariusz Bojarski, Davide Del~Testa, Daniel Dworakowski, Bernhard Firner, Beat Flepp, Prasoon Goyal, Lawrence~D Jackel, Mathew Monfort, Urs Muller, Jiakai Zhang, et~al.
\newblock End to end learning for self-driving cars.
\newblock {\em arXiv preprint arXiv:1604.07316}, 2016.

\bibitem{chen2022contrastive}
Dian Chen, Dequan Wang, Trevor Darrell, and Sayna Ebrahimi.
\newblock Contrastive test-time adaptation.
\newblock In {\em Proceedings of the IEEE/CVF Conference on Computer Vision and Pattern Recognition}, pages 295--305, 2022.

\bibitem{chen2022self}
Weijie Chen, Luojun Lin, Shicai Yang, Di Xie, Shiliang Pu, and Yueting Zhuang.
\newblock Self-supervised noisy label learning for source-free unsupervised domain adaptation.
\newblock In {\em 2022 IEEE/RSJ International Conference on Intelligent Robots and Systems (IROS)}, pages 10185--10192. IEEE, 2022.

\bibitem{de2021continual}
Matthias De~Lange, Rahaf Aljundi, Marc Masana, Sarah Parisot, Xu Jia, Ale{\v{s}} Leonardis, Gregory Slabaugh, and Tinne Tuytelaars.
\newblock A continual learning survey: Defying forgetting in classification tasks.
\newblock {\em IEEE transactions on pattern analysis and machine intelligence}, 44(7):3366--3385, 2021.

\bibitem{devlin2018bert}
Jacob Devlin, Ming-Wei Chang, Kenton Lee, and Kristina Toutanova.
\newblock Bert: Pre-training of deep bidirectional transformers for language understanding.
\newblock {\em arXiv preprint arXiv:1810.04805}, 2018.

\bibitem{dosovitskiy2017carla}
Alexey Dosovitskiy, German Ros, Felipe Codevilla, Antonio Lopez, and Vladlen Koltun.
\newblock Carla: An open urban driving simulator.
\newblock In {\em Conference on robot learning}, pages 1--16. PMLR, 2017.

\bibitem{dou2019domain}
Qi Dou, Daniel Coelho~de Castro, Konstantinos Kamnitsas, and Ben Glocker.
\newblock Domain generalization via model-agnostic learning of semantic features.
\newblock {\em Advances in Neural Information Processing Systems}, 32, 2019.

\bibitem{fernando2013unsupervised}
Basura Fernando, Amaury Habrard, Marc Sebban, and Tinne Tuytelaars.
\newblock Unsupervised visual domain adaptation using subspace alignment.
\newblock In {\em Proceedings of the IEEE international conference on computer vision}, pages 2960--2967, 2013.

\bibitem{gan2016learning}
Chuang Gan, Tianbao Yang, and Boqing Gong.
\newblock Learning attributes equals multi-source domain generalization.
\newblock In {\em Proceedings of the IEEE conference on computer vision and pattern recognition}, pages 87--97, 2016.

\bibitem{ganin2015unsupervised}
Yaroslav Ganin and Victor Lempitsky.
\newblock Unsupervised domain adaptation by backpropagation.
\newblock In {\em International conference on machine learning}, pages 1180--1189. PMLR, 2015.

\bibitem{ghifary2016scatter}
Muhammad Ghifary, David Balduzzi, W~Bastiaan Kleijn, and Mengjie Zhang.
\newblock Scatter component analysis: A unified framework for domain adaptation and domain generalization.
\newblock {\em IEEE transactions on pattern analysis and machine intelligence}, 39(7):1414--1430, 2016.

\bibitem{ghifary2015domain}
Muhammad Ghifary, W~Bastiaan Kleijn, Mengjie Zhang, and David Balduzzi.
\newblock Domain generalization for object recognition with multi-task autoencoders.
\newblock In {\em Proceedings of the IEEE international conference on computer vision}, pages 2551--2559, 2015.

\bibitem{gopalan2013unsupervised}
Raghuraman Gopalan, Ruonan Li, and Rama Chellappa.
\newblock Unsupervised adaptation across domain shifts by generating intermediate data representations.
\newblock {\em IEEE transactions on pattern analysis and machine intelligence}, 36(11):2288--2302, 2013.

\bibitem{hastie2009elements}
Trevor Hastie, Robert Tibshirani, Jerome~H Friedman, and Jerome~H Friedman.
\newblock {\em The elements of statistical learning: data mining, inference, and prediction}, volume~2.
\newblock Springer, 2009.

\bibitem{he2016deep}
K. He, X. Zhang, S. Ren, and J. Sun.
\newblock Deep residual learning for image recognition.
\newblock In {\em 2016 IEEE Conference on Computer Vision and Pattern Recognition (CVPR)}, pages 770--778, Los Alamitos, CA, USA, jun 2016. IEEE Computer Society.

\bibitem{hinton2015distilling}
Geoffrey Hinton, Oriol Vinyals, and Jeff Dean.
\newblock Distilling the knowledge in a neural network.
\newblock {\em arXiv preprint arXiv:1503.02531}, 2015.

\bibitem{hull1994database}
Jonathan~J. Hull.
\newblock A database for handwritten text recognition research.
\newblock {\em IEEE Transactions on pattern analysis and machine intelligence}, 16(5):550--554, 1994.

\bibitem{kim2019nlnl}
Youngdong Kim, Junho Yim, Juseung Yun, and Junmo Kim.
\newblock Nlnl: Negative learning for noisy labels.
\newblock In {\em Proceedings of the IEEE/CVF international conference on computer vision}, pages 101--110, 2019.

\bibitem{krizhevsky2017imagenet}
Alex Krizhevsky, Ilya Sutskever, and Geoffrey~E Hinton.
\newblock Imagenet classification with deep convolutional neural networks.
\newblock {\em Communications of the ACM}, 60(6):84--90, 2017.

\bibitem{lecun1998gradient}
Yann LeCun, L{\'e}on Bottou, Yoshua Bengio, and Patrick Haffner.
\newblock Gradient-based learning applied to document recognition.
\newblock {\em Proceedings of the IEEE}, 86(11):2278--2324, 1998.

\bibitem{lee2013pseudo}
Dong-Hyun Lee et~al.
\newblock Pseudo-label: The simple and efficient semi-supervised learning method for deep neural networks.
\newblock In {\em Workshop on challenges in representation learning, ICML}, volume~3, page 896, 2013.

\bibitem{li2018learning}
Da Li, Yongxin Yang, Yi-Zhe Song, and Timothy Hospedales.
\newblock Learning to generalize: Meta-learning for domain generalization.
\newblock In {\em Proceedings of the AAAI conference on artificial intelligence}, volume~32, 2018.

\bibitem{li2017deeper}
Da Li, Yongxin Yang, Yi-Zhe Song, and Timothy~M Hospedales.
\newblock Deeper, broader and artier domain generalization.
\newblock In {\em Proceedings of the IEEE international conference on computer vision}, pages 5542--5550, 2017.

\bibitem{li2019episodic}
Da Li, Jianshu Zhang, Yongxin Yang, Cong Liu, Yi-Zhe Song, and Timothy~M Hospedales.
\newblock Episodic training for domain generalization.
\newblock In {\em Proceedings of the IEEE/CVF International Conference on Computer Vision}, pages 1446--1455, 2019.

\bibitem{li2018domain}
Haoliang Li, Sinno~Jialin Pan, Shiqi Wang, and Alex~C Kot.
\newblock Domain generalization with adversarial feature learning.
\newblock In {\em Proceedings of the IEEE conference on computer vision and pattern recognition}, pages 5400--5409, 2018.

\bibitem{li2021progressive}
Lei Li, Ke Gao, Juan Cao, Ziyao Huang, Yepeng Weng, Xiaoyue Mi, Zhengze Yu, Xiaoya Li, and Boyang Xia.
\newblock Progressive domain expansion network for single domain generalization.
\newblock In {\em Proceedings of the IEEE/CVF Conference on Computer Vision and Pattern Recognition}, pages 224--233, 2021.

\bibitem{li2020model}
Rui Li, Qianfen Jiao, Wenming Cao, Hau-San Wong, and Si Wu.
\newblock Model adaptation: Unsupervised domain adaptation without source data.
\newblock In {\em Proceedings of the IEEE/CVF conference on computer vision and pattern recognition}, pages 9641--9650, 2020.

\bibitem{li2018deep}
Ya Li, Xinmei Tian, Mingming Gong, Yajing Liu, Tongliang Liu, Kun Zhang, and Dacheng Tao.
\newblock Deep domain generalization via conditional invariant adversarial networks.
\newblock In {\em Proceedings of the European conference on computer vision (ECCV)}, pages 624--639, 2018.

\bibitem{li2017learning}
Yuncheng Li, Jianchao Yang, Yale Song, Liangliang Cao, Jiebo Luo, and Li-Jia Li.
\newblock Learning from noisy labels with distillation.
\newblock In {\em Proceedings of the IEEE international conference on computer vision}, pages 1910--1918, 2017.

\bibitem{liang2018aggregating}
Jian Liang, Ran He, Zhenan Sun, and Tieniu Tan.
\newblock Aggregating randomized clustering-promoting invariant projections for domain adaptation.
\newblock {\em IEEE transactions on pattern analysis and machine intelligence}, 41(5):1027--1042, 2018.

\bibitem{liang2020we}
Jian Liang, Dapeng Hu, and Jiashi Feng.
\newblock Do we really need to access the source data? source hypothesis transfer for unsupervised domain adaptation.
\newblock In {\em International Conference on Machine Learning}, pages 6028--6039. PMLR, 2020.

\bibitem{liang2021source}
Jian Liang, Dapeng Hu, Yunbo Wang, Ran He, and Jiashi Feng.
\newblock Source data-absent unsupervised domain adaptation through hypothesis transfer and labeling transfer.
\newblock {\em IEEE Transactions on Pattern Analysis and Machine Intelligence}, 44(11):8602--8617, 2021.

\bibitem{liu2023deja}
Chenxi Liu, Lixu Wang, Lingjuan Lyu, Chen Sun, Xiao Wang, and Qi Zhu.
\newblock Deja vu: Continual model generalization for unseen domains.
\newblock In {\em The Eleventh International Conference on Learning Representations}, 2023.

\bibitem{liu2021source}
Yuang Liu, Wei Zhang, and Jun Wang.
\newblock Source-free domain adaptation for semantic segmentation.
\newblock In {\em Proceedings of the IEEE/CVF Conference on Computer Vision and Pattern Recognition}, pages 1215--1224, 2021.

\bibitem{long2013transfer}
Mingsheng Long, Jianmin Wang, Guiguang Ding, Jiaguang Sun, and Philip~S Yu.
\newblock Transfer feature learning with joint distribution adaptation.
\newblock In {\em Proceedings of the IEEE international conference on computer vision}, pages 2200--2207, 2013.

\bibitem{motiian2017unified}
Saeid Motiian, Marco Piccirilli, Donald~A Adjeroh, and Gianfranco Doretto.
\newblock Unified deep supervised domain adaptation and generalization.
\newblock In {\em Proceedings of the IEEE international conference on computer vision}, pages 5715--5725, 2017.

\bibitem{netzer2011reading}
Yuval Netzer, Tao Wang, Adam Coates, Alessandro Bissacco, Bo Wu, and Andrew~Y Ng.
\newblock Reading digits in natural images with unsupervised feature learning.
\newblock 2011.

\bibitem{niu2022efficient}
Shuaicheng Niu, Jiaxiang Wu, Yifan Zhang, Yaofo Chen, Shijian Zheng, Peilin Zhao, and Mingkui Tan.
\newblock Efficient test-time model adaptation without forgetting.
\newblock In {\em International conference on machine learning}, pages 16888--16905. PMLR, 2022.

\bibitem{pan2010domain}
Sinno~Jialin Pan, Ivor~W Tsang, James~T Kwok, and Qiang Yang.
\newblock Domain adaptation via transfer component analysis.
\newblock {\em IEEE transactions on neural networks}, 22(2):199--210, 2010.

\bibitem{patel2015visual}
Vishal~M Patel, Raghuraman Gopalan, Ruonan Li, and Rama Chellappa.
\newblock Visual domain adaptation: A survey of recent advances.
\newblock {\em IEEE signal processing magazine}, 32(3):53--69, 2015.

\bibitem{peng2019moment}
Xingchao Peng, Qinxun Bai, Xide Xia, Zijun Huang, Kate Saenko, and Bo Wang.
\newblock Moment matching for multi-source domain adaptation.
\newblock In {\em Proceedings of the IEEE/CVF international conference on computer vision}, pages 1406--1415, 2019.

\bibitem{qiao2020learning}
Fengchun Qiao, Long Zhao, and Xi Peng.
\newblock Learning to learn single domain generalization.
\newblock In {\em Proceedings of the IEEE/CVF Conference on Computer Vision and Pattern Recognition}, pages 12556--12565, 2020.

\bibitem{rebuffi2017icarl}
Sylvestre-Alvise Rebuffi, Alexander Kolesnikov, Georg Sperl, and Christoph~H Lampert.
\newblock icarl: Incremental classifier and representation learning.
\newblock In {\em Proceedings of the IEEE conference on Computer Vision and Pattern Recognition}, pages 2001--2010, 2017.

\bibitem{rizve2021defense}
Mamshad~Nayeem Rizve, Kevin Duarte, Yogesh~S Rawat, and Mubarak Shah.
\newblock In defense of pseudo-labeling: An uncertainty-aware pseudo-label selection framework for semi-supervised learning.
\newblock {\em arXiv preprint arXiv:2101.06329}, 2021.

\bibitem{rolnick2019experience}
David Rolnick, Arun Ahuja, Jonathan Schwarz, Timothy Lillicrap, and Gregory Wayne.
\newblock Experience replay for continual learning.
\newblock {\em Advances in Neural Information Processing Systems}, 32, 2019.

\bibitem{romera2018train}
Eduardo Romera, Luis~M Bergasa, Jose~M Alvarez, and Mohan Trivedi.
\newblock Train here, deploy there: Robust segmentation in unseen domains.
\newblock In {\em 2018 IEEE Intelligent Vehicles Symposium (IV)}, pages 1828--1833. IEEE, 2018.

\bibitem{saporta2022multi}
Antoine Saporta, Arthur Douillard, Tuan-Hung Vu, Patrick P{\'e}rez, and Matthieu Cord.
\newblock Multi-head distillation for continual unsupervised domain adaptation in semantic segmentation.
\newblock In {\em Proceedings of the IEEE/CVF Conference on Computer Vision and Pattern Recognition}, pages 3751--3760, 2022.

\bibitem{shankar2018generalizing}
Shiv Shankar, Vihari Piratla, Soumen Chakrabarti, Siddhartha Chaudhuri, Preethi Jyothi, and Sunita Sarawagi.
\newblock Generalizing across domains via cross-gradient training.
\newblock {\em arXiv preprint arXiv:1804.10745}, 2018.

\bibitem{silver2002task}
Daniel~L Silver and Robert~E Mercer.
\newblock The task rehearsal method of life-long learning: Overcoming impoverished data.
\newblock In {\em Advances in Artificial Intelligence: 15th Conference of the Canadian Society for Computational Studies of Intelligence, AI 2002 Calgary, Canada, May 27--29, 2002 Proceedings 15}, pages 90--101. Springer, 2002.

\bibitem{song2022learning}
Hwanjun Song, Minseok Kim, Dongmin Park, Yooju Shin, and Jae-Gil Lee.
\newblock Learning from noisy labels with deep neural networks: A survey.
\newblock {\em IEEE Transactions on Neural Networks and Learning Systems}, 2022.

\bibitem{sun2016return}
Baochen Sun, Jiashi Feng, and Kate Saenko.
\newblock Return of frustratingly easy domain adaptation.
\newblock In {\em Proceedings of the AAAI conference on artificial intelligence}, volume~30, 2016.

\bibitem{tang2021gradient}
Shixiang Tang, Peng Su, Dapeng Chen, and Wanli Ouyang.
\newblock Gradient regularized contrastive learning for continual domain adaptation.
\newblock In {\em Proceedings of the AAAI Conference on Artificial Intelligence}, volume~35, pages 2665--2673, 2021.

\bibitem{van2015learning}
Brendan Van~Rooyen, Aditya Menon, and Robert~C Williamson.
\newblock Learning with symmetric label noise: The importance of being unhinged.
\newblock {\em Advances in neural information processing systems}, 28, 2015.

\bibitem{volpi2018generalizing}
Riccardo Volpi, Hongseok Namkoong, Ozan Sener, John~C Duchi, Vittorio Murino, and Silvio Savarese.
\newblock Generalizing to unseen domains via adversarial data augmentation.
\newblock {\em Advances in neural information processing systems}, 31, 2018.

\bibitem{wang2020tent}
Dequan Wang, Evan Shelhamer, Shaoteng Liu, Bruno Olshausen, and Trevor Darrell.
\newblock Tent: Fully test-time adaptation by entropy minimization.
\newblock {\em arXiv preprint arXiv:2006.10726}, 2020.

\bibitem{wang2022generalizing}
Jindong Wang, Cuiling Lan, Chang Liu, Yidong Ouyang, Tao Qin, Wang Lu, Yiqiang Chen, Wenjun Zeng, and Philip Yu.
\newblock Generalizing to unseen domains: A survey on domain generalization.
\newblock {\em IEEE Transactions on Knowledge and Data Engineering}, 2022.

\bibitem{wang2018deep}
Mei Wang and Weihong Deng.
\newblock Deep visual domain adaptation: A survey.
\newblock {\em Neurocomputing}, 312:135--153, 2018.

\bibitem{wang2021learning}
Zijian Wang, Yadan Luo, Ruihong Qiu, Zi Huang, and Mahsa Baktashmotlagh.
\newblock Learning to diversify for single domain generalization.
\newblock In {\em Proceedings of the IEEE/CVF International Conference on Computer Vision}, pages 834--843, 2021.

\bibitem{yeh2021sofa}
Hao-Wei Yeh, Baoyao Yang, Pong~C Yuen, and Tatsuya Harada.
\newblock Sofa: Source-data-free feature alignment for unsupervised domain adaptation.
\newblock In {\em Proceedings of the IEEE/CVF Winter Conference on Applications of Computer Vision}, pages 474--483, 2021.

\bibitem{zenke2017continual}
Friedemann Zenke, Ben Poole, and Surya Ganguli.
\newblock Continual learning through synaptic intelligence.
\newblock In {\em International conference on machine learning}, pages 3987--3995. PMLR, 2017.

\bibitem{zhang2020generalizing}
Ling Zhang, Xiaosong Wang, Dong Yang, Thomas Sanford, Stephanie Harmon, Baris Turkbey, Bradford~J Wood, Holger Roth, Andriy Myronenko, Daguang Xu, et~al.
\newblock Generalizing deep learning for medical image segmentation to unseen domains via deep stacked transformation.
\newblock {\em IEEE transactions on medical imaging}, 39(7):2531--2540, 2020.

\bibitem{zhao2020maximum}
Long Zhao, Ting Liu, Xi Peng, and Dimitris Metaxas.
\newblock Maximum-entropy adversarial data augmentation for improved generalization and robustness.
\newblock {\em Advances in Neural Information Processing Systems}, 33:14435--14447, 2020.

\bibitem{zhou2020deep}
Kaiyang Zhou, Yongxin Yang, Timothy Hospedales, and Tao Xiang.
\newblock Deep domain-adversarial image generation for domain generalisation.
\newblock In {\em Proceedings of the AAAI Conference on Artificial Intelligence}, volume~34, pages 13025--13032, 2020.

\end{thebibliography}
}

\clearpage
\newpage

\includepdf[pages=1]{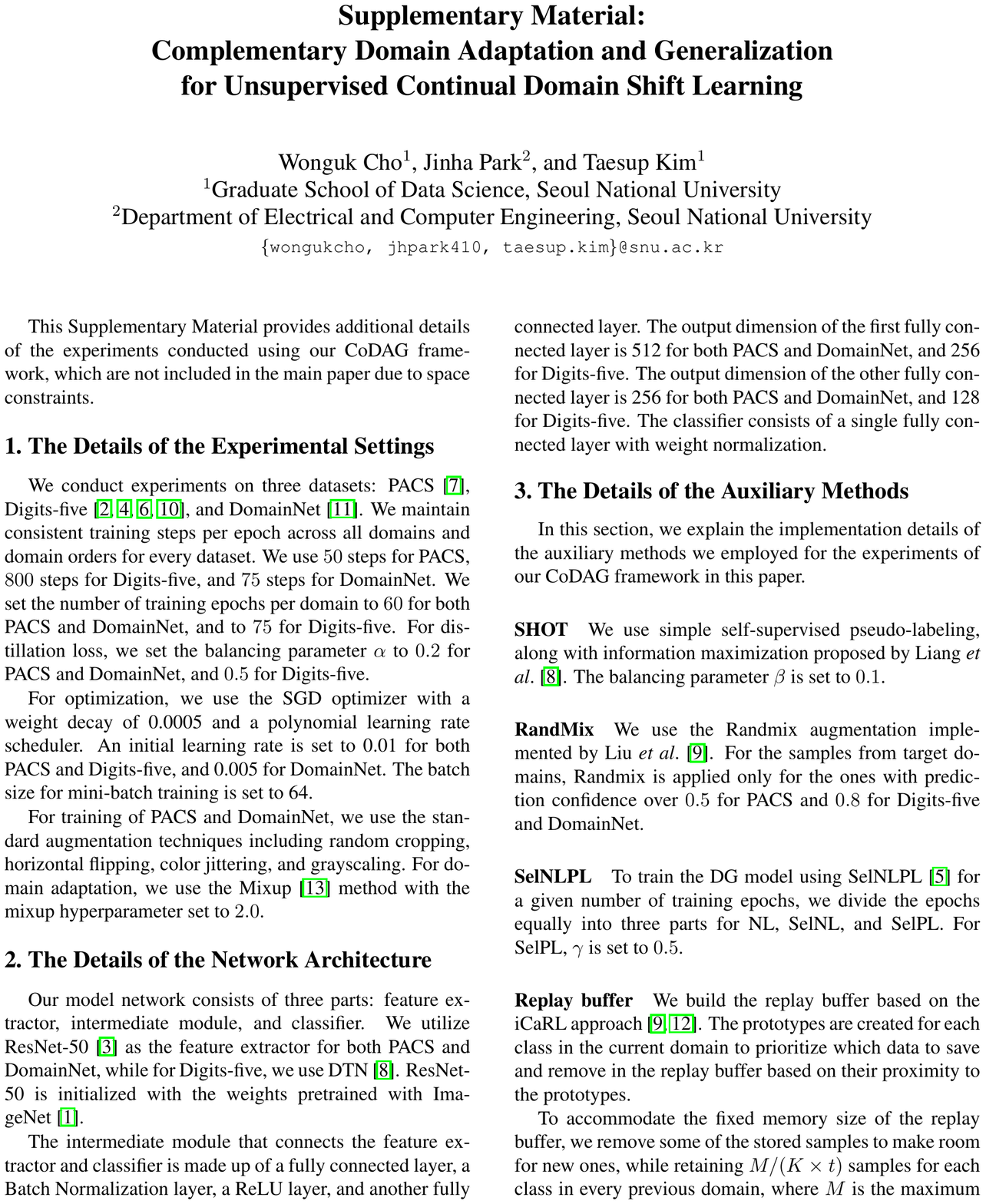}
\includepdf[pages=2]{supplementary.pdf}
\includepdf[pages=3]{supplementary.pdf}
\includepdf[pages=4]{supplementary.pdf}
\includepdf[pages=5]{supplementary.pdf}

\end{document}